\documentclass[aip,graphicx]{revtex4-1}

\usepackage{graphicx}% Include figure files
\usepackage{dcolumn}% Align imeable columns on decimal point
\usepackage{bm}% bold math

\usepackage[utf8]{inputenc}
\usepackage[T1]{fontenc}
\usepackage{mathptmx}
\usepackage{etoolbox}
\usepackage{amsmath}
\usepackage{amssymb}

\draft % marks overfull lines with a black rule on the right

\begin{document}

%\preprint{}

\title{Model-free inference of unseen attractors: Reconstructing phase space features from a single noisy trajectory using reservoir computing} 

%The Devil you don't



\author{André Röhm}
\email[]{roehm@g.ecc.u-tokyo.ac.jp}
%\thanks{}
\affiliation{Instituto de F\'{i}sica Interdisciplinar y Sistemas Complejos, IFISC (CSIC-UIB), Campus Universitat Illes Balears, E-07122 Palma de Mallorca, Spain}
\affiliation{Department of Information Physics and Computing, Graduate School of Information Science and Technology, The
University of Tokyo, 7-3-1 Hongo, Bunkyo-ku, Tokyo 113-8656, Japan}
%\affiliation{}
\author{Daniel J. Gauthier}
\affiliation{The Ohio State University, Department of Physics, 191 West Woodruff Ave., Columbus, OH 43210, USA}
\affiliation{ResCon Technologies, LLC, PO Box 21229, Columbus, OH 43221, USA}
\author{Ingo Fischer}
\email[]{ingo@ifisc.uib-csic.es}
\affiliation{Instituto de F\'{i}sica Interdisciplinar y Sistemas Complejos, IFISC (CSIC-UIB), Campus Universitat Illes Balears, E-07122 Palma de Mallorca, Spain}
%\affiliation{}


\date{\today}

\begin{abstract}
Reservoir computers are powerful tools for chaotic time series prediction. They can be trained to approximate phase space flows and can thus both predict future values to a high accuracy, as well as reconstruct the general properties of a chaotic attractor without requiring a model. In this work, we show that the ability to learn the dynamics of a complex system can be extended to systems with multiple co-existing attractors, here a 4-dimensional extension of the well-known Lorenz chaotic system. We demonstrate that a reservoir computer can infer entirely unexplored parts of the phase space: a properly trained reservoir computer can predict the existence of attractors that were never approached during training and therefore are labelled as \textit{unseen}. We provide examples where attractor inference is achieved after training solely on a single noisy trajectory.
\end{abstract}


\maketitle %\maketitle must follow title, authors, abstract and \pacs

%Lead paragraph
{
\bf
Reservoir computing is a brain-inspired machine learning scheme that can be used to mimic dynamical systems.
Reservoir computers can be trained to learn the characteristics of a target dynamical system purely from a sample time series. 
In particular, properly trained autonomous reservoir computers can act as surrogate systems while still preserving many properties of the original ground truth such as the largest Lyapunov-exponents, embedded unstable periodic orbits, or correlation measures.
Importantly, dynamical systems can exhibit more than just a single stable long-term behavior, called an attractor. 
A common scenario is the existence of pairs of symmetric solutions, but more complex co-existences can also often be found.
Systems with multiple attractors are called multistable. 
Here, we provide examples where a reservoir computer is able to learn the various attractors of a multistable system. 
We feed the reservoir just a single noisy trajectory of one of the attractors, while the other attractors remain outside of the training data range.
Then, in separate autonomous operation, the trained reservoir is able to reproduce and therefore infer the existence and shape of these \textit{unseen} attractors.
}

%In this work, we show that a trained autonomous reservoir computer's similarity to the ground truth dynamical system extends much further: 
%We use a 4-dimensional extension of the Lorenz-system that exhibits multistability as a training system. 
%We provide examples in which the reservoir computer correctly predicts the existence of previously unseen attractors, \textit{i.e.}, those without samples in the training data. 
%The reservoir computer does this without any knowledge of the target system in a model-free setting, and predicts the existence of these unseen attractors solely based on its training on a single noisy trajectory. }

\section{Introduction}

Reservoir computing is a brain-inspired machine learning technique that was popularized by the works of Maass \textit{et al.}\cite{maassRealTimeComputingStable2002} and Jaeger \cite{jaegerEchoStateApproach2001} in the form of liquid state machines and echo state networks, respectively. At its core, a reservoir computer usually consists of three elements: a fixed input layer, a fixed dynamical system with a high-dimensional phase space, called the `reservoir,' and a linear, trainable readout layer. In particular, the reservoir can be a recurrent artificial neural network (RNN) with fixed connection weights, in which case the system is still referred to as an echo state network.
% onto which the input is mapped 

Regardless of the type, the reservoir introduces transformations and memory for the input sequence by virtue of responding to an input while being in a state induced by the previous input.  
%being perpetually transient in response to the driving signals. 
By reading out the many degrees-of-freedom of the reservoir and potentially performing low-order polynomial transformations, the reservoir provides a large number of transformations of the input sequence, \textit{i.e.}, a multitude of responses. 
As opposed to other RNN schemes, the readout layer is the only part that is trained in reservoir computing. 
% For FF NN, the extreme learning machine approach also only trains the readout weights.
A linear readout layer is often sufficient, given a nonlinear reservoir, and can easily be optimized via a straightforward regression on a set of training data. 
It has recently been shown mathematically that linear reservoirs with nonlinear readouts %which were initially explored in some experiments,\cite{vandoorneExperimentalDemonstrationReservoir2014}, 
can also provide universal computational properties.\cite{gononReservoirComputingUniversality2020, gauthierNextGenerationReservoir2021a} 

It was soon discovered that reservoir computers are highly capable at predicting chaotic time series, improving accuracy by 
%`a factor of 2400' 
orders of magnitude compared to previous methods. \cite{jaegerHarnessingNonlinearityPredicting2004} 
By feeding the reservoir a chaotic time series and choosing the target to be the next point of the chaotic trajectory, the reservoir computer learns to perform so-called one-step-ahead prediction. 
Intriguingly, after training for one-step-ahead prediction, such a reservoir can be put into a `closed loop configuration,' where its \textit{own} prediction can be used as the next input. 
In this way, the reservoir computer becomes an autonomous dynamical system that can accurately continue the original data. 
%This is sometimes referred to as the `weather prediction' property.
Generally, in a chaotic attractor, the predicted time series does not agree with the specific original training data forever, as even tiny differences will eventually lead to divergence of trajectories.
However, the autonomously created time series can reproduce the general structure of the chaotic attractor it was trained on even for long time scales.
This is sometimes referred to as the `climate prediction' property, as opposed to the short-term prediction error, which analogously is referred to as 'weather prediction' property. 
It is worth emphasizing that the reservoir is reproducing the source system in a model-free manner, \textit{i.e.}, without any knowledge about the origin of the training data. 

Chaotic time series prediction using reservoir computing continues to be an active field of research with several noteworthy results. 
Pathak \textit{et al.}\cite{pathakUsingMachineLearning2017} showed that properly trained reservoir computers can be used to reconstruct the largest Lyapunov eigenvalues inside the chaotic attractor. 
Chen \textit{et al.} \cite{chenMappingTopologicalCharacteristics2020} showed that such reservoir computers also reproduce several typical geometric metrics.
Estebanez \textit{et al.}\cite{estebanezConstructiveRoleNoise2019} demonstrated how noise can be used to improve the long-term attractor reconstruction and Zhu \textit{et al.}\cite{zhuDetectingUnstablePeriodic2019} showed how delayed-feedback control can be combined with reservoir computing to find the unstable periodic orbits embedded inside the chaotic attractor. 
Furthermore, unstable fixed points can be found using reservoir computing, even when the trajectory never visits them during training. \cite{krishnagopalEncodingChaoticAttractor2019, gauthierNextGenerationReservoir2021a} 
Kim \textit{et al.}\cite{kimTeachingRecurrentNeural2021} recently showed that a reservoir computer can correctly predict bifurcations if the training is done with the bifurcation parameter as an input explicitly fed to the reservoir. 
Remarkably, they only used training trajectories taken from time series data generated below the bifurcation point to teach the reservoir the influence of the bifurcation parameter. 
This all hints at the ability of a properly trained reservoir computer 
to reproduce the phase space structure of the original system accurately and even its control parameter dependence when in autonomous mode. 

We demonstrate that the ability of phase space reconstruction can be exploited even further. 
We show that a properly trained reservoir computer cannot only reconstruct the attractor it is trained on - but it can even infer other unseen co-existing attractors in phase space and reconstruct their structure. 
By unseen co-existing attractors we specifically mean those attractors in a multistable system, whose basin of attraction was never reached during training so that no direct traces of them can be found in the original time series data.
%This property has seemingly been overlooked so far, as most of the literature has focused on the lorenz63 and Rössler system. 
To this end, we test the ability of a reservoir computer to infer unseen attractors for a four-dimensional chaotic extension of the Lorenz system with co-existing attractors in two scenarios. The first scenario comprises a torus solution co-existing with a pair of symmetric limit cycles. In the second scenario a pair of symmetric chaotic attractors coexists with a torus.

Attractor reconstruction is arguably related to the field of nonlinear system identification, which aims to derive the governing equations from a sample set of time series data. 
In particular, techniques such as Sparse Identification of Nonlinear Dynamics (SINDy),  \cite{bruntonDiscoveringGoverningEquations2016} ResNet-based approaches, \cite{qinDataDrivenGoverning2019} auto-regressive moving average (ARMA)  models and variants (ARMAx and NARMAx) are capable of producing accurate model descriptions in some cases.\cite{billingsNonlinearSystemIdentification2013}  
However, all of these either require previous knowledge about the general structure of the system or multiple sets of training trajectories. Moreover, those that only estimate the vector field would have to be combined with an integrator to facilitate one-step-ahead prediction.

In contrast, we show that reservoir computers, operated in autonomous mode, can serve as  model-free surrogates of target systems even outside of their training region with minimal input, \textit{i.e.}, they can learn to reconstruct unseen attractors after learning from a single, noisy trajectory.
Thus, reservoir computers can satisfactorily mimic target systems in cases where the training data is noisy, not a lot of training data is available, and nothing concrete about the shape of the underlying differential equation is known. 
Here, the reservoir computers do not learn the governing equations of the original system: Instead, they learn how to integrate and propagate along trajectories. Thus, the reservoir computers are learning the phase space flow without formulating any intermediate model. 
From recent studies it is known that attractor reconstruction can sometimes fail even for parameter-sets optimized for one-step-ahead prediction. \cite{haluszczynskiGoodBadPredictions2019, haluszczynskiReducingNetworkSize2020}
This also applies to the cases presented here, where a reservoir computer infers the existence of other attractors. 
Lacking a way to verify the prediction on the original system, this failure mode is harder to detect because the original data does not contain any sampling from the unseen attractors. 
Obtaining a quantitative estimate of the quality of the attractor reconstruction without knowing the target system remains an open problem.

In the following, we first introduce the model used and the target system chosen. We then analyze the reservoir computers' ability to reproduce unseen attractors. We find that for a set of parameters the reservoir computer succeeds in predicting the existence of a torus and symmetric limit cycle with low errors. In a second scenario, involving two chaotic attractors and a torus, it is more difficult to succeed. We analyze cases of partial success and discuss what the current limitations are.

\section{Theory}
\subsection{Echo State Network}

We use a continuous time version of an echo state network based on ordinary differential equations similar to those used by Lu \textit{et al.} \cite{luAttractorReconstructionMachine2018}
The reservoir computer consists of a network of $N$ coupled real-valued nodes with the total state $X \in \mathbb{R}^N$ describing all nodes evolving according to
\begin{align}
	\dot{X} &= - X + \tanh{ \left( W_{\textrm{res}} X + G W_{\textrm{in}} u(t)  + B \right) },  \label{eq:ODE_ESN}
\end{align}
where the random matrices $W_{\textrm{res}} \in \mathbb{R}^{N \times N}$ and $W_{\textrm{in}} \in \mathbb{R}^{N \times U}$ describe the internal and the input weights, respectively. The input gain is given by scalar $G$. The time-varying input $u(t) \in \mathbb{R}^{U}$ is a step-function with period $\theta$, with the amplitude of each plateau corresponding to one point of data of the discrete source series $u_k \in \mathbb{R}^{U \times K}$. 
$B \in \mathbb{R}^{N}$ is a random bias vector. The network and input matrices are sparse and the details of the random initialization and the list of all parameters are given in Appendix~\ref{app:ESN_initiliazation}. 

The state $X$ of the network is periodically recorded at intervals $\theta$ and used to construct the state matrix $S \in \mathbb{R}^{K \times N+1}$. 
We add a bias column into the state matrix such that a row $S_k$ is finally given by
\begin{align}
	S_{k} &= \lbrack X_1(t_k), X_2(t_k), \dots, X_N(t_k), 1 \rbrack,  \label{eq:rows_of_S}
\end{align}
where $t_k = T_w + k \theta  $ with $T_w$ being the `washout' or `warmup' time 
%is "washout time" a commonly used expression?
used to remove the influence of the starting state of the reservoir. 
The prediction $Y \in \mathbb{R}^{K \times U}$ of the reservoir is then given by
\begin{align}
	Y &= S W_{\textrm{out}}. \label{eq:training_of_W_out}
\end{align}
A training series of $K_{\mathrm{training}}$ elements is used to drive the system to generate $S_{\mathrm{training}}$, which, in turn, is used to determine the output weights $W_{\textrm{out}} \in \mathbb{R}^{N + 1 \times U}$. 
The error between the prediction $Y$ and the known true targets $\hat{Y}$ are minimized using an $L_2$-norm. 
To reduce the amount of over-fitting, we use Tikhonov regularization as described in Appendix~\ref{app:RC_technical_details}. The regularization strength $\eta$ is an important parameter as it controls the level of detail the training data is approximated to. High $\eta$ will fit the training data too crudely, rendering attractor inference impossible. Conversely, low $\eta$ will lead to over-fitting and poor generalization. 

Once the reservoir computer is successfully trained, we use it to probe other parts of the phase space of the target system. 
To this end, we feed the reservoir computer the beginning of a different ground-truth transient, which for now is required. 
We then observe how it evolves in autonomous mode. See Appendix~\ref{app:RC_technical_details} for details.

\subsection{Target system by Li and Sprott}

%Previous work on attractor reconstruction concentrated mainly on the chaotic Lorenz and R\"ossler attractors, as well as the Mackey-Glass delay-differential system. 
%These systems are standard examples for chaotic attractors in the dynamic systems literature and this is likely why they were chosen to explore novel reservoir computing techniques. 
%However, these systems notably lack a major feature of many complex systems: multistability of non-trivial attractors. 
%Only the Lorenz63 attractor contains a symmetric solution, which is usually suppressed and not seen as a target. 

We study the properties of unseen attractors and the ability of a reservoir computer to infer their existence. 
As our target dynamical system, we use a $4$-dimensional extension of the Lorenz attractor as proposed by Li and Sprott, \cite{liCoexistingHiddenAttractors2014} based on earlier work by Gao and Zhang,\cite{gaoNovelHyperchaoticSystem2011} whose dynamics is given by
\begin{align}
\dot{x} &= y - x + \sigma \xi_x, \label{eq:LS_x}\\
\dot{y} &= - x z + u + \sigma \xi_y, \label{eq:LS_y}\\
\dot{z} &= x y - a + \sigma \xi_z, \label{eq:LS_z}\\
\dot{u} &= - b y + \sigma \xi_u,  \label{eq:LS_u}
\end{align}
which has only two parameters $a$ and $b$. 
Here, we add additive independent identically distributed Gaussian noise terms $\xi_i$ with mean $0$ and unit variance, $E\lbrack \xi_i (t) \xi_j (t') \rbrack = \delta_{ij}  \delta_{tt'}$.
For sake of simplicity, the different variables share the same noise strength $\sigma$. 
This simplification is acceptable because all variables share the same order of magnitude. 
For $\sigma = 0$, the system is noise-free and deterministic.
We chose this system because it shows coexisting attractors of different complexities depending on parameters. 
This allows us to test the ability of reservoir computers to infer the existence of unseen attractors in various scenarios involving periodic, quasiperiodic and chaotic attractors.

By initializing the system \eqref{eq:LS_x}-\eqref{eq:LS_u} in different initial conditions, we can reach different attractors, as each trajectory will eventually reach one of the stable attractors. 
The volume of phase space from which trajectories lead to a certain attractor is called its basin of attraction.
Notably, the system \eqref{eq:LS_x}-\eqref{eq:LS_u} never features any fixed points, and hence the attractors are called hidden, and the basins of all attractors studied in this work are fractal.\cite{liCoexistingHiddenAttractors2014}

\subsection{Error estimates}

We use a quantitative measure for the quality of attractor reconstruction. 
For the reconstructed and target attractors, we take the time average for each variable $\langle x \rangle$, $\langle y \rangle$, $\langle z \rangle$, $\langle u \rangle$, and the time average of the absolute values $\langle \lvert x \lvert \rangle$, $\langle \lvert y \lvert \rangle$, $\langle \lvert z \lvert \rangle$, $\langle \lvert u \rvert \rangle$. 
The absolute values help differentiate when averages become $0$, such as for periodic states centred around the origin. 

We determine the differences $\Delta$ between prediction $x$ and the ground-truth $\Tilde{x}$, normalizing by the average ground-truth absolute averages
\begin{align}
    \Delta_x = \frac{\langle x \rangle - \langle \Tilde{x} \rangle}{\langle \lvert \Tilde{x} \lvert \rangle} \\
    \Delta_{\lvert x \lvert} = \frac{\langle \lvert x \rvert \rangle - \langle \lvert \Tilde{x} \lvert \rangle}{\langle \lvert \Tilde{x} \lvert \rangle}
\end{align}
and similarly for $y$, $z$, and $u$. 
From this, we calculate an error estimate $\Delta_{\mathrm{att}}$ for the target attractor as the root of the sum of all squares
%    \Delta_{\mathrm{total}} = \sqrt{\Delta_x^2 + \Delta_y^2 + \Delta_z^2 + \Delta_u^2 + \Delta_{\langle \lvert x \lvert \rangle}^2 + \Delta_{\langle \lvert y \lvert \rangle}^2 + \Delta_{\langle \lvert z \lvert \rangle}^2 + \Delta_{\langle \lvert u \lvert \rangle}^2} 
\begin{align}
    \Delta_{\mathrm{att}} = \sqrt{\sum_{i = \{x, y, z, u\}} \Delta_i^2 + \Delta_{\lvert i \rvert}^2 } \label{eq:delta_att}.
\end{align}
While the quantitative error $\Delta_{\mathrm{att}}$ does not capture the full picture, it allows us to easily discriminate between (partially) successful and failed attractor inference. To judge the power of a particular reservoir computer for a given scenario, we sum $\Delta_{\mathrm{att}}$ for all existing target attractors to the total error $\Delta_{\mathrm{tot}}$ via 
\begin{align}
    \Delta_{\mathrm{tot}} = \sqrt{\sum \Delta_{\mathrm{att}}^2 } .
\end{align}
The applicability of other geometric error measures can also be considered, in particular the symmetric Hausdorff distance and the average Euclidean distances per point between the real and inferred attractor. However, the former is unsuited because it only measures the worst point. The latter suffers from requiring too many data points. Our sets sample the chaotic attractors too thinly, poorly covering rare trajectories, whose points in turn dominated the average Euclidean distance. In contrast, the simple error measure $\Delta_{\mathrm{att}}$ of Eq.~\eqref{eq:delta_att} uses averages and is quite robust, even for small data set sizes.

We optimize the meta-parameters of the reservoir computers with the help of this error measure $\Delta_{\mathrm{att}}$. 
In this context, a meta-parameter is any parameter of the reservoir, \textit{i.e.}, a parameter that does not refer to the Li-Sprott oscillatory system. 
Some meta-parameters differ between simulations for the chaotic versus the limit cycle scenarios.
We obtain optimal values via a rough grid search for the number of nodes $N = 300$, the network sparsity $\rho = 0.1$, input strength $G = 0.3$ or $G=0.01$, the regularization strength $\eta = 10^{-3}$ or $\eta = 10^{-5}$, the bias strength $b = 1.0$ or $b = 3.0$, the time per input $\theta = 2.5$, the washout time $T_w = 2500$ and the maximum eigenvalue of the reservoir $\textrm{Re}(\lambda)_{\textrm{max}} = 0.95$ or $\textrm{Re}(\lambda)_{\textrm{max}} = 0.99$.

\section{Results}

\begin{figure}[htb]
	\centering \includegraphics[width=255pt]{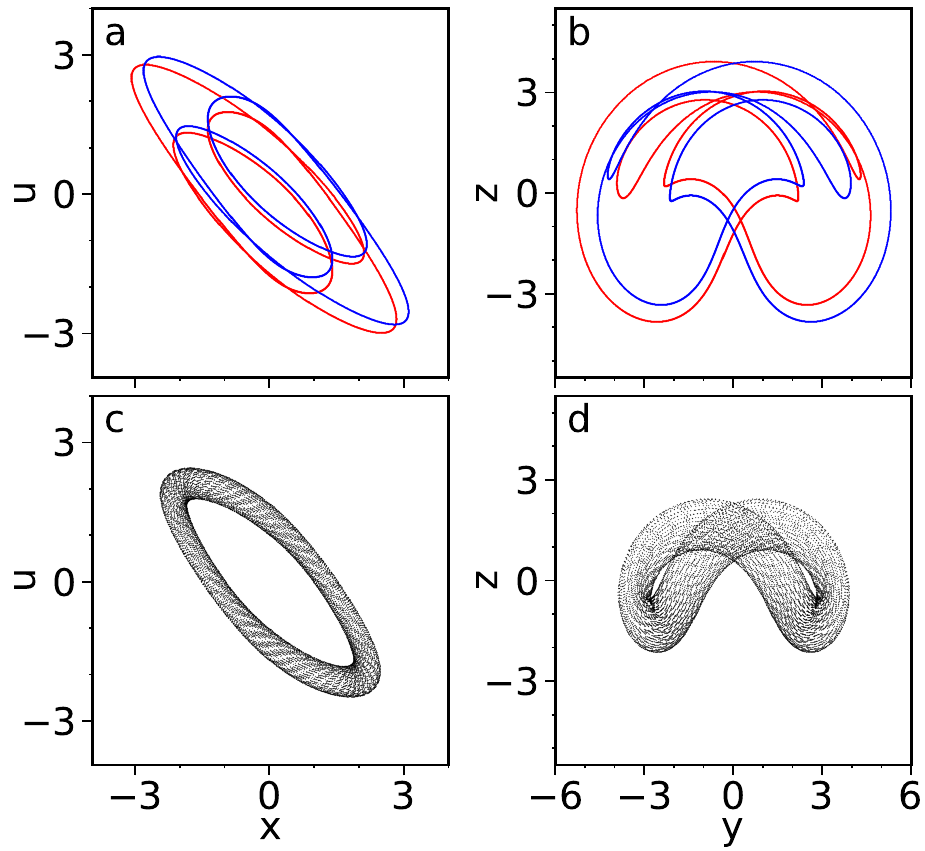}
	\caption{Solutions of the Li-Sprott equation system  for $a=2.0$, $b=0.8$ and $\sigma = 0$. Panels (a,c) show an $x$-$u$ projection, panels (b, d) a $z$-$y$ projection. A pair of symmetric limit cycles (red and blue lines in a, b) coexists with a torus (black dots in c, d). }
	\label{fig:li_sprott_solutions} 
\end{figure}

We first focus on the parameters $a=2.0$ and $b=0.8$ in the noise-free case $\sigma = 0$, for which the target system possesses three stable solutions: two periodic limit cycles and a quasiperiodic torus as shown in Fig.~\ref{fig:li_sprott_solutions}. 
The two limit cycles are symmetric with respect to the transformation $(-x, -y, z, -u)$. 
%The basin of attraction of all these solutions is fractal. 
For these parameters, the limit cycles have oscillations on a time scale of $15$ units, while the slow oscillation in the torus lasts $30$ and the fast one $5$ units, respectively. We sample the training time series with a step size of $0.3$ or about $15$ points per fast oscillation of the torus. 

The starting values for the transients used for training and details of numerical integration of the source system are described in Appendix~\ref{app:numerical_integration}. We always train on trajectories with $11,000$ data points that stay within one basin of attraction. Of these,  $1,000$ points are used as initial warm-up time $T_w$.

We then train a continuous time echo state network of Eq.~\eqref{eq:ODE_ESN} with $N=300$ on the Li-Sprott system with the phase space and attractors as shown in Fig.~\ref{fig:li_sprott_solutions}. 
We designate one of the two limit cycles as our training attractor and initialize a transient that converges to this limit cycle. 
Using this transient, we drive the reservoir computer, recording the activation of all the nodes. 
As per standard approach for attractor reconstruction, our training target is a one-step ahead prediction. 
The optimal output matrix $W_{\textrm{out}}$ is found and used subsequently. 
%Note, that during this training step, the system behaves exactly like any other attractor reconstruction system. In particular, the training data time series never leaves the attractor basin of our training attractor.

%During testing we feed the reservoir one transient for each of the other attractors of the original Li-Sprott system and observe the error. This only serves as a sanity check, to filter out solutions that drastically failed in the training.

%Since the reservoir was trained to predict the future of its input time series, it implicitly is trying to predict its own input. 
% Why implicitly?
%Closing the loop in this way is the typical manner in which a reservoir computer can be used to "mimic" a dynamical system. 
%Indeed, it has been shown that such a scheme can reproduce a wide variety of properties of the target system. 

For our goal of having a system that can find unseen attractors, we use the autonomous operation of the reservoir computer. 
For this, we feed the reservoir its own prediction. 
In our case, additional care must be taken to avoid a training failure. First, it is typical that the beginning of the training data does not directly translate into the state matrix $S_{\mathrm{training}}$.
In fact, we want the state $X$ of our reservoir to be only dependent on the drive signal $u(t)$. 
Thus, to avoid any influence of the initial state $X(0)$ of the reservoir on the training, the first $T_w$ of $u(t)$ is used to `wash out' or `warm up' the reservoir. 
We use $1,000$ of the $11,000$ training data points for this.
But the training limit cycle is highly attracting. 
Therefore, any reasonably long `wash-out' period $T_w$ leads to the loss of all information contained in the initial transient of the Li-Sprott system. 
The reservoir will effectively be trained only on the target limit cycle itself.

Second, when the training data is too low dimensional, the reservoir does not get to `see' the full shape of the dynamical flow. Even for relatively short $T_w$, the remaining training data only consists of the 1-dimensional limit cycle. 
Even when training succeeds in achieving a low fitting error on the training set, the system is susceptible to being unstable in the autonomous mode.
Because certain phase space directions might never be seen during training, the output of the reservoir computer in autonomous mode tends to diverge in those directions. 

Fortunately, these problems can be overcome with a modification that makes the training both more stable and makes the whole procedure more appropriate for realistic problems: by the inclusion of noise in the training data. 
When $\sigma > 0$, the training data produced by the Li-Sprott system \eqref{eq:LS_x}-\eqref{eq:LS_u} is in a sense permanently transient. 
Even small noise is enough to allow the trajectory to at least partially diverge from the pure limit cycle, effectively sampling all directions in the $4$-dimensional phase space. 
While not explored in detail, there is likely a trade off between training length, noise strength and regularization that influences the performance of the reservoir.
%\textbf{!!But small noise implies that we almost only see the linear stability properties and require high accuracy to identify nonlinear properties.!!}
We ensure that the noise does not lead to any attractor hopping in the training data set by visual inspection. 
We also tried using additive noise combined with a noise-free time series as has been studied previously\cite{estebanezConstructiveRoleNoise2019}, but did not achieve success, likely due the fact that we are not only targeting chaotic attractors.

In comparison to Kim \textit{et al.}, \cite{kimTeachingRecurrentNeural2021} we do not use multiple parallel training trajectories. 
Instead we only train on a single, noisy trajectory. Also, compared to their work, we use a much coarser sampling of the original time-series as they used every point of the source data without any sampling and even included the Runge-Kutta auxiliary terms. 

We always feed the reservoir all $4$ variables of the Li-Sprott system both during training and the start of the autonomous mode, \textit{i.e.}, in this work there are no unseen degrees of freedom. 
Therefore, the reservoir obtains the full information about the target system in each time step.
In principle, this should imply that no memory is needed and that removing the recurrent connections by setting $W_{res} = 0$ in an approach similar to an extreme learning machine could suffice.\cite{huangExtremeLearningMachine2004}
We performed a few preliminary investigations with $W_{res} = 0$ and find that performance worsened.
However, a detailed study of such an approach is outside the scope of this work.
In general, we believe that the results presented in this manuscript can be extended to partially observed target systems in which case memory would be necessary.
Furthermore, our chosen architecture is compatible with experimental reservoir computing, where the memory of the reservoir is an important aspect. 

\subsection*{Inferring the torus and symmetric limit cycle}

\begin{figure}[htb]
	\centering \includegraphics[width=255pt]{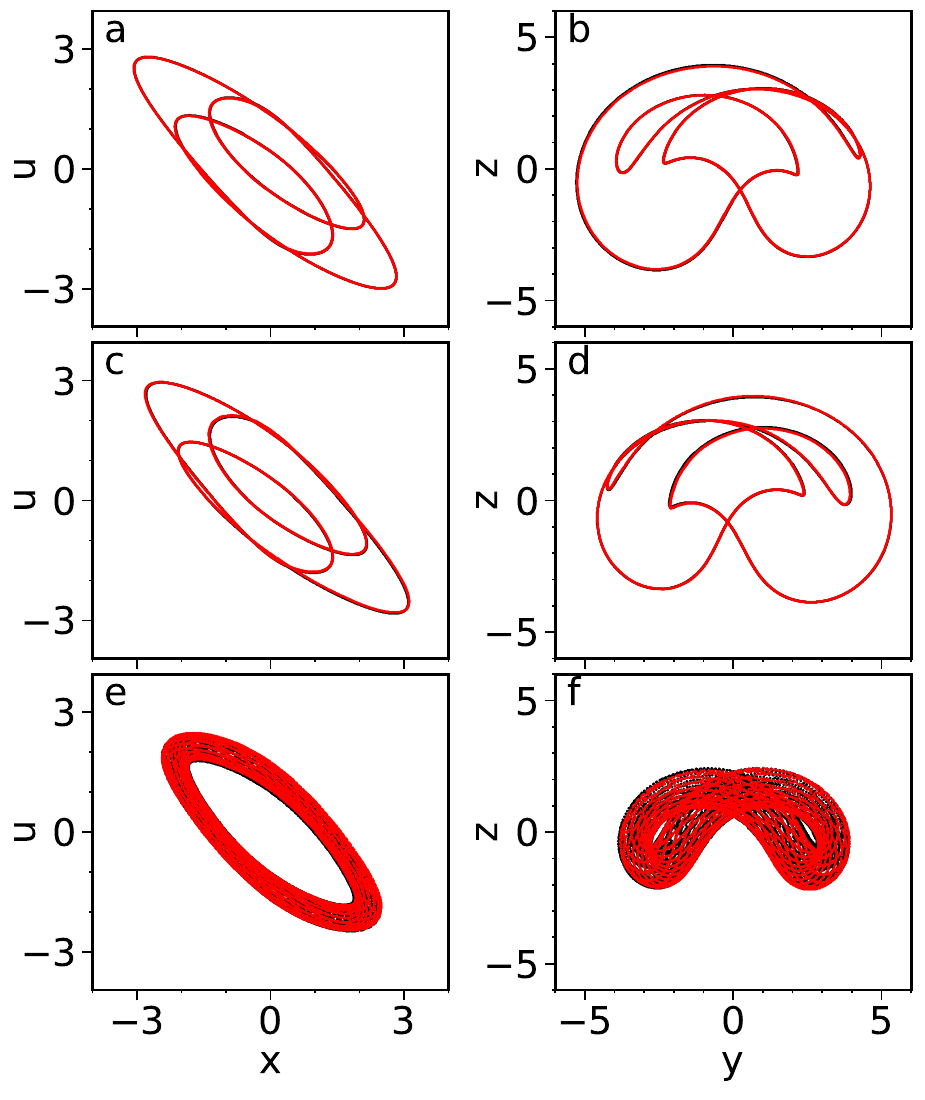}
	\caption{ An example of successful attractor inference for $a=2.0$ and $b=0.8$ of the Li-Sprott system (Eq.~\eqref{eq:LS_x}-\eqref{eq:LS_u}). The two columns show two projections of the $4$-dimensional system. The original system has three attractors (black dots): a pair of symmetric limit cycles (a-d) and a torus (e, f). A reservoir computer in autonomous mode is used to reproduce the shape ("climate") of these attractors (red points). The training is done only on the first limit cycle (a, b). The existence and shape of the other two attractors (c-f) is inferred by the reservoir. $G = 0.3$, $b = 1.0$, $\eta = 1e-3$, $\textrm{Re}(\lambda)_{\textrm{max}} = 0.95$, $\theta = 2.5$. (a, b) $\Delta_{\mathrm{att}} = 8.4 \times 10^{-3}$, (c, d) $ \Delta_{\mathrm{att}} = 7.6 \times 10^{-3}$, (e, f) $ \Delta_{\mathrm{att}} = 5.6 \times 10^{-2}$. }
	\label{fig:li_sprott_LC_success_1} 
\end{figure}

Figure \ref{fig:li_sprott_LC_success_1} shows a case of successful training and attractor inference: The system is trained on a single noisy trajectory of with $K_{training} = 10^4$ points and noise strength $\sigma \simeq 6 \times 10^{-3}$. 
For testing, we feed the system the first $1,000$ points of a ground-truth reference trajectory (see Appendix~\ref{app:RC_technical_details} for details) for each of the three attractors (top, middle and bottom row in Fig.~\ref{fig:li_sprott_LC_success_1}) and then put it into autonomous mode generating $10,000$ points on its own. 
We always use noise free ($\sigma = 0$) transients for initializing the reservoir in the autonomous mode and as targets. 

The red dots in Fig.~\ref{fig:li_sprott_LC_success_1} represent the reconstructed attractor by the reservoir, while the black dots show the true attractors of the Li-Sprott system, the latter are almost completely covered by the former. 
The system learns the attractor it is trained on (Fig.~\ref{fig:li_sprott_LC_success_1} a, b) with a low error of $\Delta_{\mathrm{att}} = 8.4 \times 10^{-3}$.
Moreover, it can also predict with high fidelity the existence and full shape of the second, symmetric limit cycle ($ \Delta_{\mathrm{att}} = 7.6 \times 10^{-3}$) and  of the torus (Fig.~\ref{fig:li_sprott_LC_success_1} e, f, $ \Delta_{\mathrm{att}} = 5.6 \times 10^{-2}$). The total error in this case was $\Delta_{\mathrm{tot}} \approx 0.06 $. 
% Do we include an RMS measure to quantify the deviation? I would be in favor (IF).
% And what about including a discussion of the involved time scales, also when discussing the limitations of our approach?

All attractors have complex shapes. 
It had been shown that a reservoir computer trained on a chaotic system can implicitly learn the position of surrounding and defining \textit{unstable} fixed points and the implicit unstable periodic orbits within. \cite{zhuDetectingUnstablePeriodic2019, krishnagopalEncodingChaoticAttractor2019, gauthierNextGenerationReservoir2021a}
Figure~\ref{fig:li_sprott_LC_success_1} clearly shows a case where we can go beyond this by demonstrating how an autonomous reservoir computer can infer entirely separate attractors in a target system. 
In particular, because the attractors of an autonomous system are always disjoint, this means that the system has to infer the properties of parts of the phase space that were neither part of the training data nor are connected to it via any trajectories, unlike unstable structures which might be connected via heteroclines. 
Here, the reservoir computer must make a plausible inference of the global phase space flow of the dynamical system from local knowledge.

It is important to note that there is quite a variation in quality of the attractor reconstruction depending on both the target as well as the randomly generated reservoir. 
In the following, we address a more difficult parameter region of the Li-Sprott system and demonstrate how partial successes and inference failure manifest themselves in such cases.

%To illustrate the variation in attractor reconstruction, we show a second set of reconstructed attractors for the same targets in Fig.~\ref{fig:li_sprott_LC_success_2}.  
%A notable difference here is that the torus got reproduced almost perfectly (Fig.~\ref{fig:li_sprott_LC_success_2} e, f), while the second limit cycle (Fig.~\ref{fig:li_sprott_LC_success_2} c, d) shows slight deviations from the ground truth (see the difference between red and black). 
%The only difference between Fig.~\ref{fig:li_sprott_LC_success_1} and Fig.~\ref{fig:li_sprott_LC_success_2} was a small change in the gain $G$ ($0.3$ from $0.1$), even the randomly created network matrices and the noisy training trajectory were identical. 
%This illustrates, that the quality of reconstruction is not a global property, with errors in one prediction not being tied to the errors of another. The total error was $\Delta \approx 0.014$, mostly driven by the deviation on the second limit cycle. 

%Some echo state networks simply fail and do not reproduce the target attractors. This includes cases, where the autonomous reservoir goes to infinity, or other cases where the wrong type of dynamics will be predicted, \textit{e.g.} a limit cycle instead of a torus or a fixed point instead of a limit cycle. This is a known issue, and we discuss it further at the end of the manuscript.
% Shall we refer to similar closed-loop prediction problems for attractor reconstruction? Ott, Dan's work, our work,.. 

\subsection*{Training in chaos and quasiperiodic dynamics}

Here, we change the target of the training. 
Using the Li-Sprott system with  $a=6.0$ and $b=0.1$, we enter a regime where two symmetric chaotic attractors coexist with a torus. 
Once again, we train the system on one of the attractors using a single noisy trajectory. 
We find that attractor reconstruction and inference in this regime are significantly harder.
In particular, we never achieve a total error $\Delta_{\mathrm{tot}} < 2 $, which is almost two orders-of-magnitude larger than in the previous case. 

\begin{figure}[htb]
	\centering \includegraphics[width=255pt]{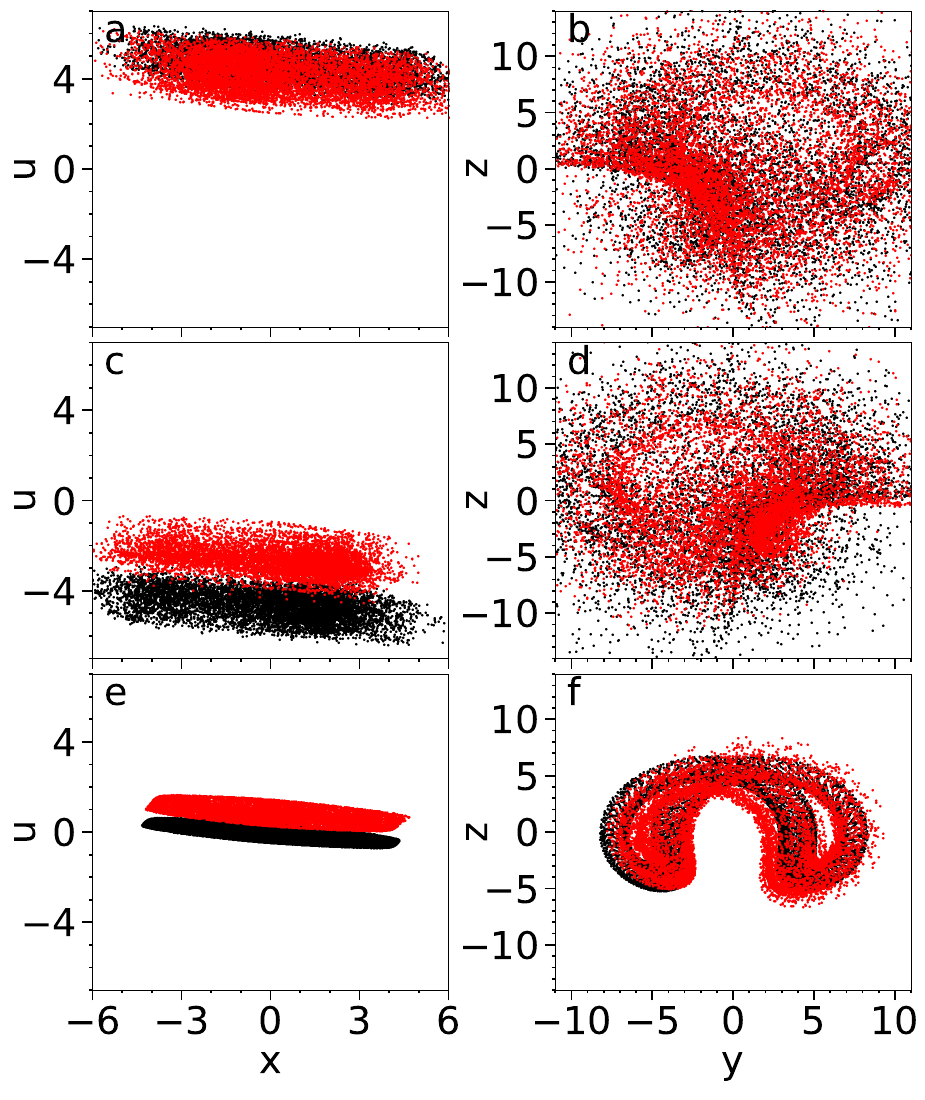}
	\caption{ Partially successful inference for $a=6.0$ and $b=0.1$. The two columns show two projections of the $4$-dimensional system. The original system has three attractors (black dots): a pair of symmetric chaotic regions (a-d) and a torus (e, f). A reservoir computer in autonomous mode is used to reproduce these attractors (red points). The training is done only on the positive-$u$ chaotic region (a, b). The existence and shape of the other chaotic attractor (c-d) and torus (e-f) is inferred by the reservoir. $G = 0.01$, $b = 3.0$, $\eta = 1e-5$, $\textrm{Re}(\lambda)_{\textrm{max}} = 0.99$, $\theta = 2.5$. (a, b) $\Delta_{\mathrm{att}} = 0.14$, (c, d) $ \Delta_{\mathrm{att}} = 0.65$, (e, f) $ \Delta_{\mathrm{att}} = 2.3$. }
	\label{fig:li_sprott_torus_success} 
\end{figure}

When training on one of the chaotic attractors, the system has to infer the existence of its symmetric counterpart and a torus. Despite extensive numerical investigation, we do not find a case where the reservoir computer fully succeeds.  
Figure~\ref{fig:li_sprott_torus_success} shows one of the most successful results with an error of $\Delta_{\mathrm{tot}} = 2.4$: the system is able to learn the chaotic region it is trained on (Fig.~\ref{fig:li_sprott_torus_success} a, b, $\Delta_{\mathrm{att}} = 0.14$) and it correctly infers the existence and type of the two other attractors, as depicted in Fig.~\ref{fig:li_sprott_torus_success} c-f. 
These other reconstructed attractors show obvious deviations. For the reconstructed symmetric chaotic region, the dynamics of the $x$, $y,$ and $z$ variables are, in fact, almost accurately predicted with errors $\Delta_i$ at most $0.18$. 
However, the $u$ variable shows a persistent positive off-set from the ground truth, as can be seen in Fig.~\ref{fig:li_sprott_torus_success} c, e, and this is reflected by higher errors $\Delta_u \approx - \Delta_{\lvert u \rvert} \approx 0.43$. 
Similarly, the torus (Fig.~\ref{fig:li_sprott_torus_success} e, f) is mismatched in the $u$-dimension with $\Delta_u \approx 0.7$. In addition, the shape is distorted increasing the error of the average absolute values $\Delta_{\lvert x \rvert} \approx 0.6$,  $\Delta_{\lvert y \rvert} \approx 1.0$, $\Delta_{\lvert z \rvert} \approx 1.9$ and $\Delta_{\lvert u \rvert} \approx - 0.3$. 

The deviation in the $u$ variable for the unseen attractors is likely due to the large separation between the training and target attractor: the two chaotic regions are clearly separated from the torus and from each other in the $u$-variable. 
We assume that this large separation makes an accurate inference harder as compared to the case of the intertwined limit cycles and torus studied earlier. The reservoir is required to correctly infer the phase space flow over longer distances, which accumulates the visible deviations between ground truth and prediction in Fig.~\ref{fig:li_sprott_torus_success} e, f. 

Furthermore, the different time scales cannot be ignored in this case. While the limit cycle and torus dynamics of the previous section allowed for a sampling that was reasonably adequate for all oscillations involved, this is not the case here. 
In particular, the chaotic dynamics show both slow oscillations on a time scale of $100$ and fast oscillations on the order of $3$. Similarly, the torus contains slow oscillations around $6$ and fast oscillations of order $120$.  This is a separation of almost two orders-of-magnitude which makes choosing a suitable sampling rate a difficult compromise. We chose a sampling interval of $0.2$, mostly adapted to the fast oscillations. 
Nevertheless, even in this more difficult case, the system correctly predicts the type, shape and rough position of the unseen attractors. If knowing the exact $u$ position and torus shape is not required, the reservoir successfully predicts the unseen attractors.

We find many cases of partial success and failure. Two types of obvious failures in particular are detectable even in a model-free setting. 
First, if the reservoir fails to reproduce the training data set, it is safe to assume that its predictions cannot be considered dependable. 
We see this type of failure remarkably often, and the fact that some randomly generated reservoir topologies work worse than others is a known issue. \cite{haluszczynskiGoodBadPredictions2019} 
A second detectable failure shows the attractor inference time-series veering off towards infinity or settle at unreasonably large values. Reservoirs with such unphysical predictions are also easily discarded.

To quantify these errors rates, we investigate the variability of the reservoir performance. 
Keeping all parameters and training targets fixed at the values used for Fig.~\ref{fig:li_sprott_torus_success}, we simulate a set of $2000$ random reservoir topologies.
Out of these $2000$ simulations over $1300$ show a deviation of $\Delta_i \geq 100$ in at least one variable. 
This indicates that the trajectory failed to converge.
Out of the remaining roughly $700$ reservoirs, only $240$ stay below an error of $\lvert \Delta_i \rvert < 2$ in every metric. These reservoirs show different degrees of success. 

\begin{figure}[htb]
	\centering \includegraphics[width=255pt]{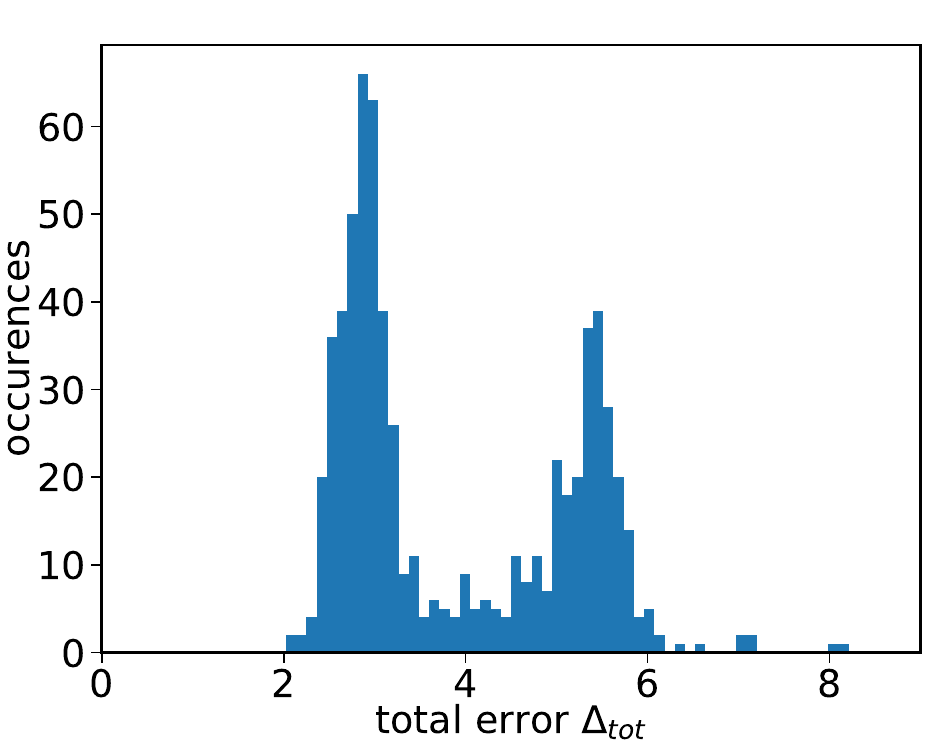}
	\caption{ Histogram of the distribution of the total error $\Delta_{\mathrm{tot}}$ for a set of $2000$ randomly generated reservoir topologies, while the other parameters are fixed. Note, that more than $1300$ simulations exhibit large errors $\Delta_{\mathrm{tot}}$ and lie outside of the range depicted here. }
	\label{fig:histogram_of_delta} 
\end{figure}

We show the subset of simulations with the lowest total error $\Delta_{\mathrm{tot}}$ in Fig.~\ref{fig:histogram_of_delta}.  The histogram shows a clear bimodal distribution. While the different predictions do not cleanly fall into classes, after investigating the source of this bimodality, we conjecture that the first peak is related to predictions that approximate the targets as good as possible. The second peak corresponds to cases in which inference fails and only the training region is predicted to be stable, \textit{i.e.}, the multistability can not be inferred. 

Ultimately, the error measures $\Delta$ always compress information. When there is a significant deviation between ground truth and prediction, many cases of partial success arise with different qualities. We show an additional example to highlight how partial success can manifest itself.

\begin{figure}[htb]
	\centering \includegraphics[width=255pt]{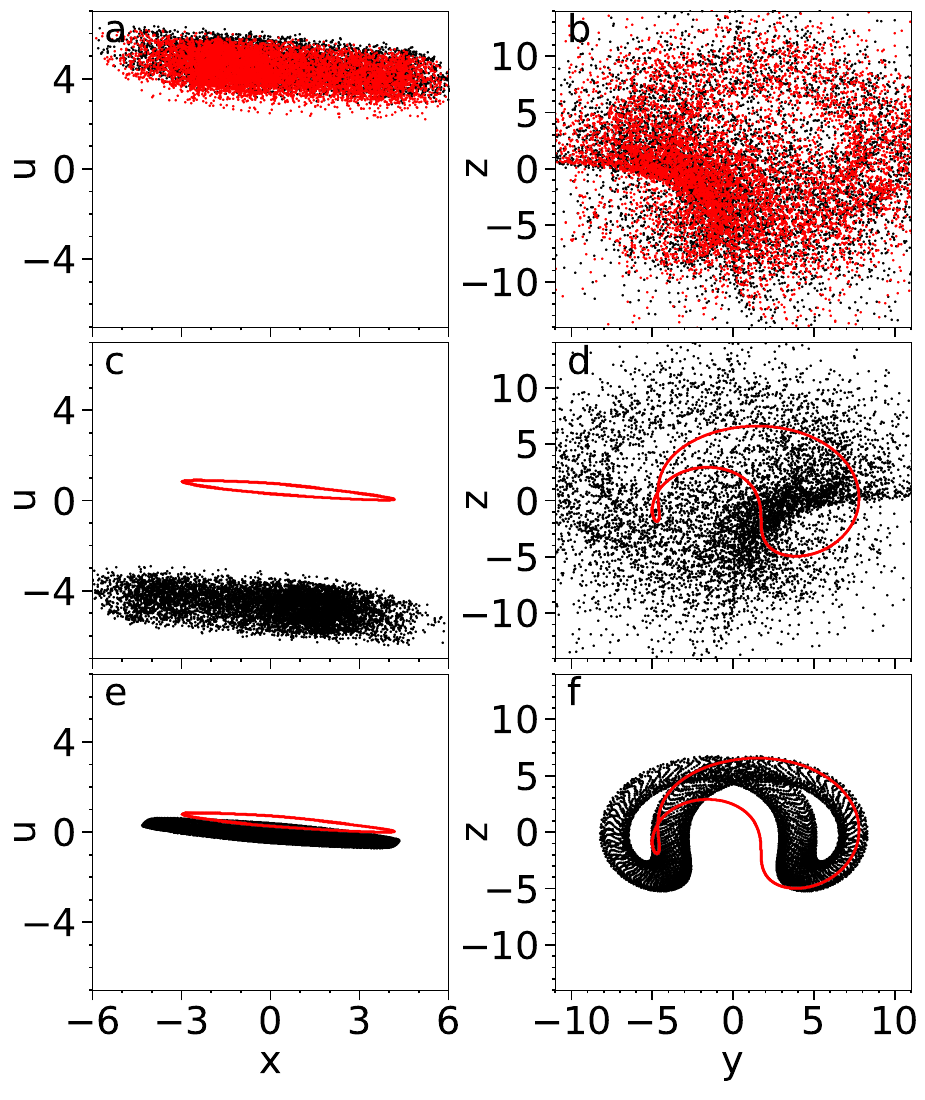}
	\caption{Partial inference success and failure for $a=6.0$ and $b=0.1$. The two columns show two projections of the $4$-dimensional system. The original system has three attractors (black dots): a pair of symmetric chaotic regions (a-d) and a torus (e, f). A reservoir computer in autonomous mode is used to reproduce these attractors (red points). The training is done only on the positive-$u$ chaotic region (a, b). The existence and shape of the other chaotic attractor (c-d) is not correctly inferred by the reservoir, despite success on the training data and instead of the torus (e-f) a limit cycle is predicted. $G = 0.01$, $b = 0.1$, $\eta = 1e-5$, $\textrm{Re}(\lambda)_{\textrm{max}} = 0.99$, $\theta = 7.5$. (a, b) $\Delta_{\mathrm{att}} = 4.5 \times 10^{-2} $, (c, d) $ \Delta_{\mathrm{att}} = 1.4$, (e, f) $ \Delta_{\mathrm{att}} = 2.1$. }
	\label{fig:li_sprott_chaos_failure} 
\end{figure}

Figure~\ref{fig:li_sprott_chaos_failure} shows an example where inference deviates from the ground truth by one attractor exhibiting a different type and another not being detected. Figure~\ref{fig:li_sprott_chaos_failure} and Fig.~\ref{fig:li_sprott_torus_success} only differ in their randomly generated topology. 
As in the successful trial, the reservoir succeeds in reproducing the training attractor (Fig.~\ref{fig:li_sprott_chaos_failure} a, b) with a low error of $\Delta_{\mathrm{att}} = 0.09 $ and therefore, barring additional information, both the attractor inference of Fig.~\ref{fig:li_sprott_chaos_failure} and Fig.~\ref{fig:li_sprott_torus_success} appear equally valid.
Nevertheless, the prediction of the two unseen attractors is less successful for Fig.~\ref{fig:li_sprott_chaos_failure}: The total error is slightly larger with $\Delta_{\mathrm{tot}} \approx 2.5$. 
Inference of the torus is not fully achieved and instead a limit cycle is predicted (Fig.~\ref{fig:li_sprott_chaos_failure} e, f, $ \Delta_{\mathrm{att}} = 2.1$). We still call this a partial success. When reconstructing the dynamic behaviour of a completely unknown dynamical system, the information that there might be an object of interest in a particular phase space region is still valuable information. The reservoir can at least predict the existence of some attracting structure despite failing to predict the torus. Similarly, we can sometimes observe cases where a torus or limit cycle is predicted with slightly distorted shape. Furthermore, this limit cycle roughly follows the outline of the torus, indicating that it correctly reflects some truth about the underlying phase-space. It might also be possible, that this limit cycle exists in the original system but is only slightly unstable. 

In contrast, the inference of the chaotic region fails for Fig.~\ref{fig:li_sprott_chaos_failure} c, d with $ \Delta_{\mathrm{att}} = 1.5$. The reservoir does not even detect the existence of an attracting structure and converges to the same attractor as Fig.~\ref{fig:li_sprott_chaos_failure} e, f. In this case, there is no obvious trace in the reconstructed reservoir time-series about this symmetrical chaotic region. 

In general, we observe that the behaviour in cases of such partial successes and failures exhibits a broad range of types. It ranges from underestimating the width of a torus over simplifications of attractor shapes to predicting the entirely wrong class of attractor, \textit{e.g.}, a limit cycle in place of the torus. 
Furthermore, the quality of reproduction of one of the unseen attractors does not directly translate to the quality of the other. 

For now, we have no method for preventing this kind of failure. 
If a method could be devised that can judge the quality of inferred attractors, one could use this to discriminate between successful and failing reservoirs. 
One potential avenue is a scenario for which one does know two attractors of the target system: In this case, one can use one of them as a training data set, and another as an independent test. For systems with only a single training data set, independent error estimation remains an open question for now.

A final open question regards the role of noise in the training data: minute details of the phase space flow that might be important for long range inference are washed out by the noise. 
We therefore tried to reduce the noise as much as possible. 
However, a certain level of noise is necessary for our scheme to succeed.
Systematically investigating the relationship between necessary noise levels (as compared to computational noise levels), training series length, inference distance and regularization strength are interesting challenges for future work to give insights into these important aspects.

%Finally, to illustrate how the system behaves in progressively more difficult circumstances, we show Figure~\ref{fig:li_sprott_torus_failure}. 
%Here, we trained the echo state network on one of the two chaotic attractors (Figure~\ref{fig:li_sprott_torus_failure} a, b). 
%The system then fails at reproducing the other symmetric chaotic attractor (Figure~\ref{fig:li_sprott_torus_failure} c, d) as well as the torus (Figure~\ref{fig:li_sprott_torus_failure} e, f). 
%Instead of the torus, a limit cycle is predicted in its place. 
%The error was $\Delta \approx 52$. 
%This failure is enlightening, in that it still shows some partial functioning. First, while the reservoir is unable to predict the correct type, it still correctly predicts the existence of another attractor in the corresponding region of the phase space (a limit cycle instead of a torus). 
%When reconstructing the dynamic behaviour of a completely unknown dynamical system, the information that there might be an object of interest in a particular phase space region is still valuable information. 
%Arguably, this limit cycle even follows the rough outline of the true torus. 
%The echo state network fails, however, to predict anything about the symmetric chaotic attractor (compare red prediction and black truth in Figure~\ref{fig:li_sprott_torus_failure} c, d). 
%The echo state network converges to the aforementioned limit cycle in the torus region instead. 
%Again, we assume that the two symmetric chaotic regions are too far separated in the $u$-variable. 

\section{Discussion}

%Global vs local learning discussion (flow definition)
%Hills vs valleys

We report the successful use of an autonomously operated reservoir computer for attractor reconstruction. 
This includes not only the reconstruction of the training attractor, but also of attractors that the system never sees during training. 
Moreover, we perform training using only a single noisy trajectory. 
This demonstrates that reservoir computers are powerful tools for model-free system analysis and attractor prediction, as there are no assumptions put into the reservoir about the target system.
% Consistent use of tenses!?

We show how the system is able to correctly predict the existence and shape of a pair of symmetric intertwined complex limit cycles and a torus in a 4-dimensional extension of the Lorenz system. 
We then explore more difficult scenarios involving two chaotic regions and a torus. 
For this problem, we obtain partial success.
And even for cases when we cannot find reservoirs that fully succeeded in attractor inference, some useful partial information is still extracted such as the approximate location of an attractor, or just the existence of `some kind' of attractor. 

Several open questions remain to further develop this approach: First, and foremost, the fraction of randomly generated reservoirs that fail is quite high. 
For non-optimized parameters, it can be more than $95 \%$. Even for optimized networks, the percentage remains high: For the parameters used in Fig.~\ref{fig:li_sprott_LC_success_1} about $1500$ out of $2000$ randomly generated reservoirs cannot predict the existence of the symmetric limit cycle solution. 
The reservoir might simply fail to predict any kind of other attractor in the system. 
Some failures are detectable even if nothing else is known about the target system. 
Some failures, however, produce plausibly looking but false reconstructions. 
Finding methods for quality control of the reproduced attractors remains an open problem.

The inference errors are higher in the second investigated case involving two chaotic regions and a torus. 
Here, we observe much larger deviations $\Delta_i$ on average. Out of $2000$ simulations differing only in their random topology, only $251$ did not exceed $\Delta_i > 2$ in at least one metric. 
Over $1300$ even had a deviation of $\Delta_i \geq 100$ in at least one variable indicating that the trajectory failed to converge.
We speculate, that this due to larger separation in phase space of the attractors, as well as the larger difference in time scales of the involved oscillations.
The widely separated time scales might be better tackled and the prediction errors alleviated by using different reservoirs adapted to the different time scales. This remains a question for future studies. 
%One could think about using more complex reservoirs, instead of the simple echo state networks used here, to potentially alleviate some of these problems. 
% That's why I also brought in the discussion of the time scales. Some suggestion would be helpful.

Another remaining question concerns the full understanding of the role of noise.\cite{billingsNonlinearSystemIdentification2013} 
In our scheme, noise in the source system helps to explore all directions in phase space and enables the system training to succeed. 
However, noise also results in a destruction of fine details. 
Finding the best trade-off between noise and accuracy remains a future challenge, which will give important hints as to the kind of information the reservoir is exploiting when inferring the existence of unseen attractors. 
Conversely, it will be fruitful to study how to achieve improved training on noise-free data or data with minimal noise.

We speculate that the Ridge regression parameter likely plays a crucial role. 
It controls how much importance the reservoir assigns to small differences in the training data, and to what extent it tries to reproduce those. 
In fact, the Ridge regression parameter might control the `model complexity': training with strong regularization should lead to simpler models, while training with weak regularization allows for more complexity, but is also more susceptible to overfitting. 
This also relates to the influence of finite precision and its role in attractor inference. 

Taking a broader view, the reservoir has to perform its task based on a finite amount of noisy data. 
Thus, the amount of information that can be extracted about `unseen parts' of the phase space should also have some fundamental limit. 
We expect that a more complex target dynamical system will also require a longer training sequence for a reservoir to be able to emulate it. 
Questions about the relationship of data length, target system complexity and related fundamental limits will give further insights about how close to an `optimal' prediction can be made by a particular reservoir computer.
%Finally, a quantitative comparison with other methods from nonlinear systems identification would be of significant interest. 

Furthermore, in a real-world scenario, a way to fully explore the phase space without the need of initial transients from other regions is vital. We suspect that a rough initialization with artificial data such as constant values is possible but have not tested this proposition.
%Similarly, when nothing is known about the target system, a model-free quantitative estimate for the quality of the inferred attractors would greatly aid further research in this direction.
% Isn't this a repetition from before??

%!! We should end on a positive note!!

With the examples shown in our work, we provide a proof-of-principle what can be achieved. A reservoir computer can definitely infer the existence of unseen attractors with varying degrees of success. As such, this further proves their suitability as model-free substitutes for a target system. 

%If one believes the previously posed assumption, then we might expect suitably trained biological brains to perform similarly or better to the reservoir computers presented here.

\begin{acknowledgments}
A.R. and I.F. acknowledge support by the Spanish State Research Agency (AEI), through the Severo Ochoa and Maria de Maeztu Program for Centers and Units of Excellence in R\&D (Grant No. MDM-2017-0711).
\end{acknowledgments}

\section*{Data Availability Statement}
The data that support the findings of this study are available from the corresponding author upon reasonable request.

\appendix

\section{Construction of the random echo state network}
\label{app:ESN_initiliazation}

The echo state networks used in this work always consists of $N=300$ real-valued nodes, whose vectorial state $X$ evolves according to 
\begin{align}
	\dot{X} &= - X + \tanh{ \left( W_{\textrm{res}} X + G W_{\textrm{in}} u(t)  + B \right) }, 
\end{align}
The real-valued reservoir weight matrix $W_{\textrm{res}}$, input weight matrix $W_{\textrm{in}}$ and bias vector $B$ are random but fixed and initialized as follows: 
The bias vector $B$ consists of $N=300$ uniformly distributed random numbers in $\lbrack - b, b \rbrack$. 
$W_{\textrm{in}}$ is a $300 \times 4$ dimensional matrix for our system. Each of the $N=300$ rows has only a single non-zero entry, whose column is randomly chosen with equal probabilities. 
The real-valued entry itself is drawn from a uniform random distribution in $\lbrack - 1, 1 \rbrack$ for each row. 
The strength of the input weights is globally controlled via the parameter $G$.

Finally, the reservoir weight matrix $W_{\textrm{res}}$ is a sparse matrix with sparsity $\rho$. 
We used $\rho=0.1$ in all presented simulations. 
Larger $\rho$ were tried but did not yield any immediately noticeable differences, and as sparse matrices are faster to simulate, we chose the smallest acceptable $\rho$. 
The non-zero values are initially distributed in $\lbrack - 1, 1 \rbrack$. 
After creation, we calculate the largest real part of the eigenvalues $ \textrm{Re} (\Lambda)_{\textrm{max}}$ of the resulting matrix $W_{\textrm{sparse}}$. We then divide the entire matrix and multiply by a margin factor, such that the resulting reservoir weight matrix $W_{\textrm{res}}$ possesses a maximal real part of any eigenvalue with $\textrm{Re}(\lambda)_{\textrm{max}}$, that are related through
\begin{align}
	W_{\textrm{res}} = \textrm{Re}(\lambda)_{\textrm{max}} \frac{W_{\textrm{sparse}}}{\textrm{Re} (\Lambda)_{\textrm{max}}} .
\end{align}
In our simulations, $\textrm{Re}(\lambda)_{\textrm{max}}$ is either $0.95$ or $0.99$, depending on what is found to be better.

\section{Numerical integration of the target system}
\label{app:numerical_integration}

The 4-dimensional extension of the Lorenz system \eqref{eq:LS_x}-\eqref{eq:LS_u} as proposed by Li and Sprott \cite{liCoexistingHiddenAttractors2014} is numerically integrated using an Euler-Maruyama integration scheme written in C++.
A high-fidelity time series is created with an integration step of $h=10^{-3}$. 
From this, we sample the input time series of the reservoir computer every $300$ steps for the first parameter regime ($a = 2.0$, $b=0.8$), and every 200 steps for the second parameter region ($a = 6.0$, $b=0.1$). 
The parameter $h$ has no connection to $\theta$, which is a reservoir computer parameter, whereas here we are solely concerned with the source time series preparation. 
This corresponds roughly to taking $15$ points per fast oscillation. 
Slower oscillations are sampled much more, with up to $500$ points in the torus or chaotic regions. 

To reach the two limit cycles for $a = 2.0$, $b=0.8$, we start the time series at $(\pm 5, \pm 1, 1 \pm1)$. 
To reach the torus solution, we start at $(4, 1, -1, 1)$. 
To reach the two chaotic attractors for $a = 6.0$, $b=0.1$, the state is initialized at $(0,  \mp 4, 0, \pm 5)$; the torus was reached from $(1, -1, 1, -1)$.
We created both trajectories with and without noise. 
The noise terms $\xi_i$ in \eqref{eq:LS_x}-\eqref{eq:LS_u} are drawn from a pseudo-random standard Gaussian distribution with mean zero and unit variance.
The noise terms in different variables are uncorrelated.
For the trajectories with noise, the standard deviation of the noise \textit{in the numerical integration} is set to $0.2$; with a time step of $h=10^{-3}$ this implies that the noise strength $\sigma$ in \eqref{eq:LS_x}-\eqref{eq:LS_u} is $0.2 / \sqrt{10^{-3}} \approx 6 \times 10^{-3}$. 
The final sampled time series $u_k$ has length $11000$, of which we use $1000$ points for washout, $9999$ points for training and $1$ point as reserve to have the future target.

\section{Reservoir computing technical details}
\label{app:RC_technical_details}

We integrate \eqref{eq:ODE_ESN} with a fourth order Runge-Kutta integrator. 
The state of all nodes $X_n \in \mathbb{R}$ is initialized as $0$ in each element. 
The integration time step is $0.1$. 
We let the echo state network evolve without input for $300$ time units or $3000$ steps, and then switch to the "warm up" or "wash out" procedure with input. 
The input sequence $u_k$ is fed as a piece-wise constant function $u(t) \in \mathbb{R}^4$ with interval lengths $\theta$. We use an interval length of $\theta = 2.5$ found through meta-parameter scans, albeit the exact length does not critically influence the performance.
Each component of $u(t)$ corresponds to one of the four variables of the Li-Sprott-variant of the Lorenz system \eqref{eq:LS_x}-\eqref{eq:LS_u}.

We use the first $1,000$ points encoded in $u(t)$ to remove any influence of the starting state, \textit{i.e.}, $T_w = 2,500$.
After that, we start recording the system state for $10^4$ inputs, \textit{i.e.}, $25,000$ time units, in accordance to \eqref{eq:rows_of_S}, \textit{i.e.}, every $\theta$ time units the state of all $300$ nodes is recorded. 
We always use the last integration point in each piece-wise constant interval of $u(t)$ to record the maximally large reaction to the input.
We obtain a state matrix $S$ with $9,999$ rows and $N + 1 = 301$ columns. 

We create a target vector $Y$ with $9,999$ rows and $4$ columns. 
Each column contains one variable of the input $u_k$ shifted one step into the future.
Because both input and output are $4$-dimensional, the output weight matrix $W_{\textrm{out}}$ is of size $(N+1) \times 4$.
We then solve the following equation for $W_{\textrm{out}}$ using the \textit{solve}-function of the \textit{armadillo} C++ wrapper \cite{sandersonArmadilloTemplatebasedLibrary} of the Open-BLAS linear algebra package
\begin{align}
     S^T Y = (S^T S + \eta I_{N+1} ) W_{\textrm{out}},
\end{align}    
where $\eta$ is the regularization factor of the Tikhonov regularization and $I_{N+1}$ the identity matrix of appropriate size $N+1$. $S^T$ is the transpose of the state matrix.

\subsubsection*{Autonomous operation}

For the autonomous operation, we use the output weights $W_{\textrm{out}}$ learned during training. 
First, we reset the system state $X$ to the point it was after the first $300$ time units of input-free evolution. 
We then feed the first $1,000$ ground truth points of a noise-free transient $u_k$ leading to one of the three target attractors. 

We observe the system state $X$ at the end of this washout-period, and construct state vector $S_i(X) \in \mathbb{R}^{301}$ that looks like recording the row of a state matrix \eqref{eq:rows_of_S}. 
But instead of saving its entries, we directly multiply with $W_{\textrm{out}}$, which yields $4$ values corresponding to the $4$ dimensions of the predicted output stream. 
We treat these $4$ values as the next element of a self-generating input sequence $\Tilde{u}_k$ and keep track of $\Tilde{u}_k$. 
With the new response $X$ generated, we repeat the process for $10,000$ steps. 
The series $\Tilde{u}_k$ then is the reconstructed attractor as inferred or learned by the echo state network. 



\begin{thebibliography}{21}%
\makeatletter
\providecommand \@ifxundefined [1]{%
 \@ifx{#1\undefined}
}%
\providecommand \@ifnum [1]{%
 \ifnum #1\expandafter \@firstoftwo
 \else \expandafter \@secondoftwo
 \fi
}%
\providecommand \@ifx [1]{%
 \ifx #1\expandafter \@firstoftwo
 \else \expandafter \@secondoftwo
 \fi
}%
\providecommand \natexlab [1]{#1}%
\providecommand \enquote  [1]{``#1''}%
\providecommand \bibnamefont  [1]{#1}%
\providecommand \bibfnamefont [1]{#1}%
\providecommand \citenamefont [1]{#1}%
\providecommand \href@noop [0]{\@secondoftwo}%
\providecommand \href [0]{\begingroup \@sanitize@url \@href}%
\providecommand \@href[1]{\@@startlink{#1}\@@href}%
\providecommand \@@href[1]{\endgroup#1\@@endlink}%
\providecommand \@sanitize@url [0]{\catcode `\\12\catcode `\$12\catcode
  `\&12\catcode `\#12\catcode `\^12\catcode `\_12\catcode `\%12\relax}%
\providecommand \@@startlink[1]{}%
\providecommand \@@endlink[0]{}%
\providecommand \url  [0]{\begingroup\@sanitize@url \@url }%
\providecommand \@url [1]{\endgroup\@href {#1}{\urlprefix }}%
\providecommand \urlprefix  [0]{URL }%
\providecommand \Eprint [0]{\href }%
\providecommand \doibase [0]{http://dx.doi.org/}%
\providecommand \selectlanguage [0]{\@gobble}%
\providecommand \bibinfo  [0]{\@secondoftwo}%
\providecommand \bibfield  [0]{\@secondoftwo}%
\providecommand \translation [1]{[#1]}%
\providecommand \BibitemOpen [0]{}%
\providecommand \bibitemStop [0]{}%
\providecommand \bibitemNoStop [0]{.\EOS\space}%
\providecommand \EOS [0]{\spacefactor3000\relax}%
\providecommand \BibitemShut  [1]{\csname bibitem#1\endcsname}%
\let\auto@bib@innerbib\@empty
%</preamble>
\bibitem [{\citenamefont {Maass}, \citenamefont {Natschl{\"a}ger},\ and\
  \citenamefont {Markram}(2002)}]{maassRealTimeComputingStable2002}%
  \BibitemOpen
  \bibfield  {author} {\bibinfo {author} {\bibfnamefont {W.}~\bibnamefont
  {Maass}}, \bibinfo {author} {\bibfnamefont {T.}~\bibnamefont
  {Natschl{\"a}ger}}, \ and\ \bibinfo {author} {\bibfnamefont {H.}~\bibnamefont
  {Markram}},\ }\bibfield  {title} {\enquote {\bibinfo {title} {Real-{{Time
  Computing Without Stable States}}: {{A New Framework}} for {{Neural
  Computation Based}} on {{Perturbations}}},}\ }\href {\doibase
  10.1162/089976602760407955} {\bibfield  {journal} {\bibinfo  {journal}
  {Neural Computation}\ }\textbf {\bibinfo {volume} {14}},\ \bibinfo {pages}
  {2531--2560} (\bibinfo {year} {2002})}\BibitemShut {NoStop}%
\bibitem [{\citenamefont {Jaeger}(2001)}]{jaegerEchoStateApproach2001}%
  \BibitemOpen
  \bibfield  {author} {\bibinfo {author} {\bibfnamefont {H.}~\bibnamefont
  {Jaeger}},\ }\bibfield  {title} {\enquote {\bibinfo {title} {The" echo state"
  approach to analysing and training recurrent neural networks-with an erratum
  note'},}\ }\href@noop {} {\bibfield  {journal} {\bibinfo  {journal} {Bonn,
  Germany: German National Research Center for Information Technology GMD
  Technical Report}\ }\textbf {\bibinfo {volume} {148}} (\bibinfo {year}
  {2001})}\BibitemShut {NoStop}%
\bibitem [{\citenamefont {Gonon}\ and\ \citenamefont
  {Ortega}(2020)}]{gononReservoirComputingUniversality2020}%
  \BibitemOpen
  \bibfield  {author} {\bibinfo {author} {\bibfnamefont {L.}~\bibnamefont
  {Gonon}}\ and\ \bibinfo {author} {\bibfnamefont {J.-P.}\ \bibnamefont
  {Ortega}},\ }\bibfield  {title} {\enquote {\bibinfo {title} {Reservoir
  {{Computing Universality With Stochastic Inputs}}},}\ }\href {\doibase
  10.1109/TNNLS.2019.2899649} {\bibfield  {journal} {\bibinfo  {journal} {IEEE
  Transactions on Neural Networks and Learning Systems}\ }\textbf {\bibinfo
  {volume} {31}},\ \bibinfo {pages} {100--112} (\bibinfo {year}
  {2020})}\BibitemShut {NoStop}%
\bibitem [{\citenamefont {Gauthier}\ \emph {et~al.}(2021)\citenamefont
  {Gauthier}, \citenamefont {Bollt}, \citenamefont {Griffith},\ and\
  \citenamefont {Barbosa}}]{gauthierNextGenerationReservoir2021a}%
  \BibitemOpen
  \bibfield  {author} {\bibinfo {author} {\bibfnamefont {D.~J.}\ \bibnamefont
  {Gauthier}}, \bibinfo {author} {\bibfnamefont {E.}~\bibnamefont {Bollt}},
  \bibinfo {author} {\bibfnamefont {A.}~\bibnamefont {Griffith}}, \ and\
  \bibinfo {author} {\bibfnamefont {W.~A.~S.}\ \bibnamefont {Barbosa}},\
  }\bibfield  {title} {\enquote {\bibinfo {title} {Next generation reservoir
  computing},}\ }\href {\doibase 10.1038/s41467-021-25801-2} {\bibfield
  {journal} {\bibinfo  {journal} {Nature Communications}\ }\textbf {\bibinfo
  {volume} {12}},\ \bibinfo {pages} {5564} (\bibinfo {year}
  {2021})}\BibitemShut {NoStop}%
\bibitem [{\citenamefont {Jaeger}\ and\ \citenamefont
  {Haas}(2004)}]{jaegerHarnessingNonlinearityPredicting2004}%
  \BibitemOpen
  \bibfield  {author} {\bibinfo {author} {\bibfnamefont {H.}~\bibnamefont
  {Jaeger}}\ and\ \bibinfo {author} {\bibfnamefont {H.}~\bibnamefont {Haas}},\
  }\bibfield  {title} {\enquote {\bibinfo {title} {Harnessing {{Nonlinearity}}:
  {{Predicting Chaotic Systems}} and {{Saving Energy}} in {{Wireless
  Communication}}},}\ }\href {\doibase 10.1126/science.1091277} {\bibfield
  {journal} {\bibinfo  {journal} {Science}\ }\textbf {\bibinfo {volume}
  {304}},\ \bibinfo {pages} {78--80} (\bibinfo {year} {2004})}\BibitemShut
  {NoStop}%
\bibitem [{\citenamefont {Pathak}\ \emph {et~al.}(2017)\citenamefont {Pathak},
  \citenamefont {Lu}, \citenamefont {Hunt}, \citenamefont {Girvan},\ and\
  \citenamefont {Ott}}]{pathakUsingMachineLearning2017}%
  \BibitemOpen
  \bibfield  {author} {\bibinfo {author} {\bibfnamefont {J.}~\bibnamefont
  {Pathak}}, \bibinfo {author} {\bibfnamefont {Z.}~\bibnamefont {Lu}}, \bibinfo
  {author} {\bibfnamefont {B.~R.}\ \bibnamefont {Hunt}}, \bibinfo {author}
  {\bibfnamefont {M.}~\bibnamefont {Girvan}}, \ and\ \bibinfo {author}
  {\bibfnamefont {E.}~\bibnamefont {Ott}},\ }\bibfield  {title} {\enquote
  {\bibinfo {title} {Using machine learning to replicate chaotic attractors and
  calculate {{Lyapunov}} exponents from data},}\ }\href {\doibase
  10.1063/1.5010300} {\bibfield  {journal} {\bibinfo  {journal} {Chaos: An
  Interdisciplinary Journal of Nonlinear Science}\ }\textbf {\bibinfo {volume}
  {27}},\ \bibinfo {pages} {121102} (\bibinfo {year} {2017})}\BibitemShut
  {NoStop}%
\bibitem [{\citenamefont {Chen}\ \emph {et~al.}(2020)\citenamefont {Chen},
  \citenamefont {Weng}, \citenamefont {Yang}, \citenamefont {Gu}, \citenamefont
  {Zhang},\ and\ \citenamefont
  {Small}}]{chenMappingTopologicalCharacteristics2020}%
  \BibitemOpen
  \bibfield  {author} {\bibinfo {author} {\bibfnamefont {X.}~\bibnamefont
  {Chen}}, \bibinfo {author} {\bibfnamefont {T.}~\bibnamefont {Weng}}, \bibinfo
  {author} {\bibfnamefont {H.}~\bibnamefont {Yang}}, \bibinfo {author}
  {\bibfnamefont {C.}~\bibnamefont {Gu}}, \bibinfo {author} {\bibfnamefont
  {J.}~\bibnamefont {Zhang}}, \ and\ \bibinfo {author} {\bibfnamefont
  {M.}~\bibnamefont {Small}},\ }\bibfield  {title} {\enquote {\bibinfo {title}
  {Mapping topological characteristics of dynamical systems into neural
  networks: {{A}} reservoir computing approach},}\ }\href {\doibase
  10.1103/PhysRevE.102.033314} {\bibfield  {journal} {\bibinfo  {journal}
  {Physical Review E}\ }\textbf {\bibinfo {volume} {102}},\ \bibinfo {pages}
  {033314} (\bibinfo {year} {2020})}\BibitemShut {NoStop}%
\bibitem [{\citenamefont {Est{\'e}banez}, \citenamefont {Fischer},\ and\
  \citenamefont {Soriano}(2019)}]{estebanezConstructiveRoleNoise2019}%
  \BibitemOpen
  \bibfield  {author} {\bibinfo {author} {\bibfnamefont {I.}~\bibnamefont
  {Est{\'e}banez}}, \bibinfo {author} {\bibfnamefont {I.}~\bibnamefont
  {Fischer}}, \ and\ \bibinfo {author} {\bibfnamefont {M.~C.}\ \bibnamefont
  {Soriano}},\ }\bibfield  {title} {\enquote {\bibinfo {title} {Constructive
  {{Role}} of {{Noise}} for {{High}}-{{Quality Replication}} of {{Chaotic
  Attractor Dynamics Using}} a {{Hardware}}-{{Based Reservoir Computer}}},}\
  }\href {\doibase 10.1103/PhysRevApplied.12.034058} {\bibfield  {journal}
  {\bibinfo  {journal} {Physical Review Applied}\ }\textbf {\bibinfo {volume}
  {12}},\ \bibinfo {pages} {034058} (\bibinfo {year} {2019})}\BibitemShut
  {NoStop}%
\bibitem [{\citenamefont {Zhu}, \citenamefont {Ma},\ and\ \citenamefont
  {Lin}(2019)}]{zhuDetectingUnstablePeriodic2019}%
  \BibitemOpen
  \bibfield  {author} {\bibinfo {author} {\bibfnamefont {Q.}~\bibnamefont
  {Zhu}}, \bibinfo {author} {\bibfnamefont {H.}~\bibnamefont {Ma}}, \ and\
  \bibinfo {author} {\bibfnamefont {W.}~\bibnamefont {Lin}},\ }\bibfield
  {title} {\enquote {\bibinfo {title} {Detecting unstable periodic orbits based
  only on time series: {{When}} adaptive delayed feedback control meets
  reservoir computing},}\ }\href {\doibase 10.1063/1.5120867} {\bibfield
  {journal} {\bibinfo  {journal} {Chaos: An Interdisciplinary Journal of
  Nonlinear Science}\ }\textbf {\bibinfo {volume} {29}},\ \bibinfo {pages}
  {093125} (\bibinfo {year} {2019})}\BibitemShut {NoStop}%
\bibitem [{\citenamefont {Krishnagopal}\ \emph {et~al.}(2019)\citenamefont
  {Krishnagopal}, \citenamefont {Katz}, \citenamefont {Girvan},\ and\
  \citenamefont {Reggia}}]{krishnagopalEncodingChaoticAttractor2019}%
  \BibitemOpen
  \bibfield  {author} {\bibinfo {author} {\bibfnamefont {S.}~\bibnamefont
  {Krishnagopal}}, \bibinfo {author} {\bibfnamefont {G.}~\bibnamefont {Katz}},
  \bibinfo {author} {\bibfnamefont {M.}~\bibnamefont {Girvan}}, \ and\ \bibinfo
  {author} {\bibfnamefont {J.}~\bibnamefont {Reggia}},\ }\bibfield  {title}
  {\enquote {\bibinfo {title} {Encoding of a {{Chaotic Attractor}} in a
  {{Reservoir Computer}}: {{A Directional Fiber Investigation}}},}\ }in\ \href
  {\doibase 10.1109/IJCNN.2019.8851853} {\emph {\bibinfo {booktitle} {2019
  {{International Joint Conference}} on {{Neural Networks}} ({{IJCNN}})}}}\
  (\bibinfo {year} {2019})\ pp.\ \bibinfo {pages} {1--8}\BibitemShut {NoStop}%
\bibitem [{\citenamefont {Kim}\ \emph {et~al.}(2021)\citenamefont {Kim},
  \citenamefont {Lu}, \citenamefont {Nozari}, \citenamefont {Pappas},\ and\
  \citenamefont {Bassett}}]{kimTeachingRecurrentNeural2021}%
  \BibitemOpen
  \bibfield  {author} {\bibinfo {author} {\bibfnamefont {J.~Z.}\ \bibnamefont
  {Kim}}, \bibinfo {author} {\bibfnamefont {Z.}~\bibnamefont {Lu}}, \bibinfo
  {author} {\bibfnamefont {E.}~\bibnamefont {Nozari}}, \bibinfo {author}
  {\bibfnamefont {G.~J.}\ \bibnamefont {Pappas}}, \ and\ \bibinfo {author}
  {\bibfnamefont {D.~S.}\ \bibnamefont {Bassett}},\ }\bibfield  {title}
  {\enquote {\bibinfo {title} {Teaching recurrent neural networks to infer
  global temporal structure from local examples},}\ }\href {\doibase
  10.1038/s42256-021-00321-2} {\bibfield  {journal} {\bibinfo  {journal}
  {Nature Machine Intelligence}\ }\textbf {\bibinfo {volume} {3}},\ \bibinfo
  {pages} {316--323} (\bibinfo {year} {2021})}\BibitemShut {NoStop}%
\bibitem [{\citenamefont {Brunton}, \citenamefont {Proctor},\ and\
  \citenamefont {Kutz}(2016)}]{bruntonDiscoveringGoverningEquations2016}%
  \BibitemOpen
  \bibfield  {author} {\bibinfo {author} {\bibfnamefont {S.~L.}\ \bibnamefont
  {Brunton}}, \bibinfo {author} {\bibfnamefont {J.~L.}\ \bibnamefont
  {Proctor}}, \ and\ \bibinfo {author} {\bibfnamefont {J.~N.}\ \bibnamefont
  {Kutz}},\ }\bibfield  {title} {\enquote {\bibinfo {title} {Discovering
  governing equations from data by sparse identification of nonlinear dynamical
  systems},}\ }\href {\doibase 10.1073/pnas.1517384113} {\bibfield  {journal}
  {\bibinfo  {journal} {Proceedings of the National Academy of Sciences}\
  }\textbf {\bibinfo {volume} {113}},\ \bibinfo {pages} {3932--3937} (\bibinfo
  {year} {2016})}\BibitemShut {NoStop}%
\bibitem [{\citenamefont {Qin}, \citenamefont {Wu},\ and\ \citenamefont
  {Xiu}(2019)}]{qinDataDrivenGoverning2019}%
  \BibitemOpen
  \bibfield  {author} {\bibinfo {author} {\bibfnamefont {T.}~\bibnamefont
  {Qin}}, \bibinfo {author} {\bibfnamefont {K.}~\bibnamefont {Wu}}, \ and\
  \bibinfo {author} {\bibfnamefont {D.}~\bibnamefont {Xiu}},\ }\bibfield
  {title} {\enquote {\bibinfo {title} {Data driven governing equations
  approximation using deep neural networks},}\ }\href {\doibase
  10.1016/j.jcp.2019.06.042} {\bibfield  {journal} {\bibinfo  {journal}
  {Journal of Computational Physics}\ }\textbf {\bibinfo {volume} {395}},\
  \bibinfo {pages} {620--635} (\bibinfo {year} {2019})}\BibitemShut {NoStop}%
\bibitem [{\citenamefont
  {Billings}(2013)}]{billingsNonlinearSystemIdentification2013}%
  \BibitemOpen
  \bibfield  {author} {\bibinfo {author} {\bibfnamefont {S.~A.}\ \bibnamefont
  {Billings}},\ }\href@noop {} {\emph {\bibinfo {title} {Nonlinear {{System
  Identification}}: {{NARMAX Methods}} in the {{Time}}, {{Frequency}}, and
  {{Spatio}}-{{Temporal Domains}}}}}\ (\bibinfo  {publisher} {{John Wiley \&
  Sons}},\ \bibinfo {year} {2013})\BibitemShut {NoStop}%
\bibitem [{\citenamefont {Haluszczynski}\ and\ \citenamefont
  {R{\"a}th}(2019)}]{haluszczynskiGoodBadPredictions2019}%
  \BibitemOpen
  \bibfield  {author} {\bibinfo {author} {\bibfnamefont {A.}~\bibnamefont
  {Haluszczynski}}\ and\ \bibinfo {author} {\bibfnamefont {C.}~\bibnamefont
  {R{\"a}th}},\ }\bibfield  {title} {\enquote {\bibinfo {title} {Good and bad
  predictions: {{Assessing}} and improving the replication of chaotic
  attractors by means of reservoir computing},}\ }\href {\doibase
  10.1063/1.5118725} {\bibfield  {journal} {\bibinfo  {journal} {Chaos: An
  Interdisciplinary Journal of Nonlinear Science}\ }\textbf {\bibinfo {volume}
  {29}},\ \bibinfo {pages} {103143} (\bibinfo {year} {2019})}\BibitemShut
  {NoStop}%
\bibitem [{\citenamefont {Haluszczynski}\ \emph {et~al.}(2020)\citenamefont
  {Haluszczynski}, \citenamefont {Aumeier}, \citenamefont {Herteux},\ and\
  \citenamefont {R{\"a}th}}]{haluszczynskiReducingNetworkSize2020}%
  \BibitemOpen
  \bibfield  {author} {\bibinfo {author} {\bibfnamefont {A.}~\bibnamefont
  {Haluszczynski}}, \bibinfo {author} {\bibfnamefont {J.}~\bibnamefont
  {Aumeier}}, \bibinfo {author} {\bibfnamefont {J.}~\bibnamefont {Herteux}}, \
  and\ \bibinfo {author} {\bibfnamefont {C.}~\bibnamefont {R{\"a}th}},\
  }\bibfield  {title} {\enquote {\bibinfo {title} {Reducing network size and
  improving prediction stability of reservoir computing},}\ }\href {\doibase
  10.1063/5.0006869} {\bibfield  {journal} {\bibinfo  {journal} {Chaos: An
  Interdisciplinary Journal of Nonlinear Science}\ }\textbf {\bibinfo {volume}
  {30}},\ \bibinfo {pages} {063136} (\bibinfo {year} {2020})}\BibitemShut
  {NoStop}%
\bibitem [{\citenamefont {Lu}, \citenamefont {Hunt},\ and\ \citenamefont
  {Ott}(2018)}]{luAttractorReconstructionMachine2018}%
  \BibitemOpen
  \bibfield  {author} {\bibinfo {author} {\bibfnamefont {Z.}~\bibnamefont
  {Lu}}, \bibinfo {author} {\bibfnamefont {B.~R.}\ \bibnamefont {Hunt}}, \ and\
  \bibinfo {author} {\bibfnamefont {E.}~\bibnamefont {Ott}},\ }\bibfield
  {title} {\enquote {\bibinfo {title} {Attractor reconstruction by machine
  learning},}\ }\href {\doibase 10.1063/1.5039508} {\bibfield  {journal}
  {\bibinfo  {journal} {Chaos: An Interdisciplinary Journal of Nonlinear
  Science}\ }\textbf {\bibinfo {volume} {28}},\ \bibinfo {pages} {061104}
  (\bibinfo {year} {2018})}\BibitemShut {NoStop}%
\bibitem [{\citenamefont {Li}\ and\ \citenamefont
  {Sprott}(2014)}]{liCoexistingHiddenAttractors2014}%
  \BibitemOpen
  \bibfield  {author} {\bibinfo {author} {\bibfnamefont {C.}~\bibnamefont
  {Li}}\ and\ \bibinfo {author} {\bibfnamefont {J.~C.}\ \bibnamefont
  {Sprott}},\ }\bibfield  {title} {\enquote {\bibinfo {title} {Coexisting
  {{Hidden Attractors}} in a 4-{{D Simplified Lorenz System}}},}\ }\href
  {\doibase 10.1142/S0218127414500345} {\bibfield  {journal} {\bibinfo
  {journal} {International Journal of Bifurcation and Chaos}\ }\textbf
  {\bibinfo {volume} {24}},\ \bibinfo {pages} {1450034} (\bibinfo {year}
  {2014})}\BibitemShut {NoStop}%
\bibitem [{\citenamefont {Gao}\ and\ \citenamefont
  {Zhang}(2011)}]{gaoNovelHyperchaoticSystem2011}%
  \BibitemOpen
  \bibfield  {author} {\bibinfo {author} {\bibfnamefont {Z.-Z.}\ \bibnamefont
  {Gao}}\ and\ \bibinfo {author} {\bibfnamefont {C.}~\bibnamefont {Zhang}},\
  }\bibfield  {title} {\enquote {\bibinfo {title} {A {{Novel Hyperchaotic
  System}}},}\ }\href@noop {} {\bibfield  {journal} {\bibinfo  {journal}
  {Journal of Jishou University (Natural Sciences Edition)}\ }\textbf {\bibinfo
  {volume} {32}},\ \bibinfo {pages} {65} (\bibinfo {year} {2011})}\BibitemShut
  {NoStop}%
\bibitem [{\citenamefont {Huang}, \citenamefont {Zhu},\ and\ \citenamefont
  {Siew}(2004)}]{huangExtremeLearningMachine2004}%
  \BibitemOpen
  \bibfield  {author} {\bibinfo {author} {\bibfnamefont {G.-B.}\ \bibnamefont
  {Huang}}, \bibinfo {author} {\bibfnamefont {Q.-Y.}\ \bibnamefont {Zhu}}, \
  and\ \bibinfo {author} {\bibfnamefont {C.-K.}\ \bibnamefont {Siew}},\
  }\bibfield  {title} {\enquote {\bibinfo {title} {Extreme learning machine: A
  new learning scheme of feedforward neural networks},}\ }in\ \href {\doibase
  10.1109/IJCNN.2004.1380068} {\emph {\bibinfo {booktitle} {2004 {{IEEE
  International Joint Conference}} on {{Neural Networks}} ({{IEEE Cat}}.
  {{No}}.{{04CH37541}})}}},\ Vol.~\bibinfo {volume} {2}\ (\bibinfo {year}
  {2004})\ pp.\ \bibinfo {pages} {985--990 vol.2}\BibitemShut {NoStop}%
\bibitem [{\citenamefont {Sanderson}\ and\ \citenamefont
  {Curtin}()}]{sandersonArmadilloTemplatebasedLibrary}%
  \BibitemOpen
  \bibfield  {author} {\bibinfo {author} {\bibfnamefont {C.}~\bibnamefont
  {Sanderson}}\ and\ \bibinfo {author} {\bibfnamefont {R.}~\bibnamefont
  {Curtin}},\ }\bibfield  {title} {\enquote {\bibinfo {title} {Armadillo: A
  template-based {{C}}++ library for linear algebra},}\ }\href@noop {} {\ ,\
  \bibinfo {pages} {7}}\BibitemShut {NoStop}%
\end{thebibliography}
\end{document}